\documentclass{article}
\usepackage{spconf,amsmath,epsfig}
\usepackage{amssymb,multirow,booktabs}

\title{Dual-Branched Spatio-temporal Fusion Network for Multi-horizon Tropical Cyclone Track Forecast}
\name{Zili Liu, Kun Hao, Xiaoyi Geng and Zhenwei Shi}
\address{Image Processing Center, School of Astronautics, Beihang University, Beijing 100191, China}
\begin{document}
%
\maketitle
\begin{abstract}


Tropical cyclone (TC) is an extreme tropical weather system and its trajectory can be described by variety of spatio-temporal data. Effective mining of these data is the key to accurate TCs track forecasting. However, existing methods face the problem that the model complexity is too high or it is difficult to efficiently extract features from multi-modal data. In this paper, we propose the Dual-Branched spatio-temporal Fusion Network (DBF-Net) -- a novel multi-horizon tropical cyclone track forecasting model which fuses the multi-modal features efficiently. DBF-Net contains a TC features branch that extracts temporal features from 1D inherent features of TCs and a pressure field branch that extracts spatio-temporal features from reanalysis 2D pressure field. Through the encoder-decoder-based architecture and efficient feature fusion, DBF-Net can fully mine the information of the two types of data, and achieve good TCs track prediction results. Extensive experiments on historical TCs track data in the Northwest Pacific show that our DBF-Net achieves significant improvement compared with existing statistical and deep learning TCs track forecast methods. 
\end{abstract}

\begin{keywords}
Tropical cyclones, tracking forecast, \\spatio-temporal data, multi-modal data fusion
\end{keywords}

\section{Introduction}
\label{sec:intro}
Tropical cyclones (TCs) are low-pressure vortexes occurring over the tropical or subtropical oceans, which are one of the major meteorological disasters facing mankind. Depending on the region of occurrence, they are often referred to as typhoons or hurricanes. Accurate forecasting for the TCs trajectory can greatly reduce the damage to people and property caused by TCs.

The research on TCs track forecast has gone through four stages since the 1960s, including empirical methods, statistical methods, numerical methods and deep learning methods. Early methods of TCs track forecast were limited by observation techniques and computational devices and could rely only on the subjective experience for achieve forecasting. Thus, some traditional methods such as extrapolation and similar path methods were developed. From 1980s, with the rapid development of statistical models, forecasting models based on statistical regression methods such as Climatology and Persistence (CLIPER) \cite{CLIPER} were proposed one after another. However, poor representation capabilities and manual feature selection make it difficult to produce accurate forecast results. Since the 1990s, due to the continuous improvement of observation techniques and computer performance, the Numerical Weather Prediction (NWP) system (e.g. American National Hurricane Center Track and Intensity Model) gradually become the mainstream choice for official meteorological forecasting agency. NWP achieves forecasting by solving complex partial differential equations of weather dynamics, which is very computationally expensive and requires the support of supercomputer platforms. In recent years, machine learning especially deep learning has developed rapidly. Various deep neural networks (DNNs) based on deep learning have shown outstanding performance in tasks such as computer vision \cite{zou2019object}, natural language processing \cite{transformer}, time series forecasting \cite{Informer}, etc. Since the computational complexity of DNNs is much smaller than that of traditional NWP models, researchers have proposed many different DNNs to predict TCs track \cite{PCT,Wang,Kordmahalleh,Alemany,GAO2018A,ConvLSTM,2019Tropical,BiGRU-attn}, which is also the research content in this paper.

\begin{figure}[t]
\centering
\includegraphics[width=0.95\columnwidth]{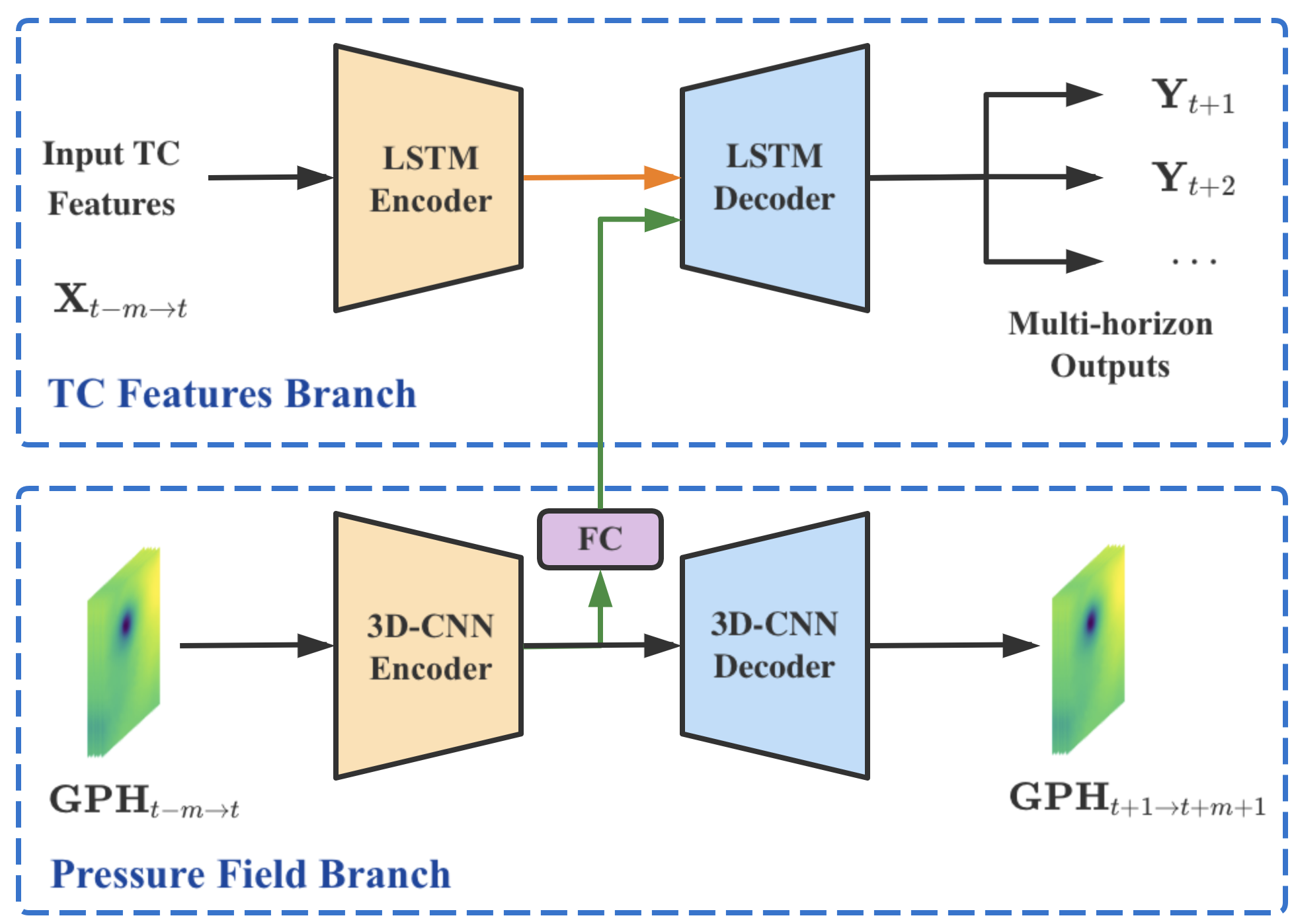} 
\caption{The overall structure of the proposed DBF-Net for multi-horizon TC track forecast.}
\label{fig1}
\end{figure}

TCs is a complex weather system, and its trajectory is affected by various physical quantities in the atmosphere, such as pressure field, wind field and so on. Therefore, different from the data used in traditional image recognition or natural language processing tasks, the data used to describe typhoon track naturally contains multi-modal temporal aligned data. In this paper, we divide these data into three categories: \emph{inherent features of TCs}, \emph{remote sensing images} and \emph{meteorological fields}. The key to deep learning based forecasting methods is, to some extent, the full exploitation of different types of data.

The \emph{inherent features of TCs} at a particular time are always represented by a column vector or tensor, which contains information such as the latitude, longitude intensity of the center of TCs at that time. Infantile deep-learning-based TCs track forecast model mainly utilize historical inherent features of TCs to predict the future locations of the TCs. The classical multi-layer perceptions (MLP) \cite{PCT,Wang} and various of time series prediction models, such as Recurrent Neural Networks (RNNs) \cite{Kordmahalleh,Alemany} , Long Short Term Memory (LSTM) model \cite{GAO2018A} and Bi-direction Gate Recurrent Unit model \cite{BiGRU-attn}, are used to learn the time series pattern of the data. However, the time series models based only on inherent features of TCs usually have low accuracy of track forecast due to the lack of consideration of factors affecting TCs' trajectory.

Compared with the one-dimensional (1D) \emph{inherent features of TCs} vector, the 2D \emph{remote sensing images} and \emph{meteorological fields} can describe the relevant information around the TCs. As for the TCs track forecasting using remote sensing images, it can be seen as a special kind of video frame prediction task. M. R{\"u}ttgers et al. uses a generative adversarial network (GAN) to predict the TCs track images and the corresponding location of TCs center \cite{GAN}. Wu et al. proposes a multitask machine learning framework based on an improved GAN to predict the track and intensity of TCs simultaneously \cite{WGAN}. The track forecast methods above takes full advantage of the powerful performance of GANs in the field of computer vision. However, the remote sensing images used in such methods need to be acquired from geostationary satellites to ensure the high temporal resolution, and the images can not represent the physical factors affecting the TCs trajectory.

\emph{meteorological fields}, such as pressure fields and wind fields, are the main factors affecting the trajectory of TCs. In 2017, M.Mudigonda et al. \cite{Seg} proposed the CNN-LSTM model for segmenting and tracking TCs and verified the direct high correlation of TC tracks in meteorological fields. S. Kim proposes a ConvLSTM-based spatio-tempral model predicts the trajectory map based on the density map sequence generated from the wind velocity and precipitation fields \cite{ConvLSTM}. But the predicted trajectory map can not reflect the exact location of the TCs precisely.  Therefore, how to efficiently fuse the \emph{meteorological fields} data into the TCs track forecast model to improve the forecast accuracy has gradually become the mainstream research direction in recent years \cite{GANS,2019Tropical}. Due to the large variation in the distribution of different meteorological fields, S. Giffard-Roisin et al. \cite{2019Tropical} uses different CNN models to encode the reanalysis data of wind and pressure field respectively, and fuse them with past track data of TCs. However, multiple CNN models increase the number of parameters and computational complexity of the forecast model, and the model is difficult to train. In addition, the inclusion of excessive use of meteorological field data weakens the role of the inherent features data of TCs and does not adequately learn the time-series features of the data. Therefore, how to efficiently utilize the meteorological field data and fully exploit the intrinsic time-series information of the inherent features of TCs still needs further research.

To solve the above problems, this paper fully exploits the temporal information in the inherent features data of TCs and the spatio-temporal information in reanalysis 2D pressure field data, and propose a \emph{Dual-Branched spatio-temporal Fusion Network} (DBF-Net) for multi-horizon tropical cyclone track forecast (i.e. predicting the TCs' track at multiple future time steps). Specifically, as shown in Fig. 1, in TC features branch, a LSTM-based encoder-decoder network is used to capture the high-level temporal features of the input TC features and provide the multi-horizon TCs trajectory forecasting outputs. In pressure field branch, a 2D-CNN-based encoder-decoder network is used to extract the spatio-temporal features from the geopotential height (GPH) around TCs that can be used to complement the track forecasting information by predicting the GPH at multiple future time steps. Besides, the high-level spatio-temporal feature obtained from the pressure field branch is fused to the LSTM-decoder in TC features branch through a fully connected layer. 

Through efficient spatio-temporal feature extraction and fusion of the two types of data, the 24h forecast accuracy of DBF-Net on historical TC tracks data in the Northwest Pacific (WNP) is 119km which is much better compared with other deep-learning based method \cite{2019Tropical,BiGRU-attn}. Besides, we also compare our work with other traditional method, such as extrapolation,  CLIPER model and NWP methods. Finally, we exhibit the forecast results for several individual cases of TC events for further analysis and verification.

\section{Methodology}
\label{sec:pagestyle}
In this section, we will explain our proposed Dual-Branched spatio-temporal Fusion Network (DBF-Net). The overall architecture of the proposed DBF-Net is shown in Fig.1. The two branches contained in DBF-Net will be split into three sub-modules, and introduced separately. Algorithmic details about the DBF-Net will be mentioned in the last sub-section of this section.

\subsection{Preliminaries}
We formally introduce symbols and notations in this sub-section. In DBF-Net, there are two types of data as the input, where  $\mathcal{X}_t=\{\textbf{X}_i\in \mathbb{R}^p\ |\ i\in [t-m,t], i\in \mathbb{Z}\}$ represents the input historical inherent features sequence of TC and $\mathcal{G}_{t}=\{\textbf{GPH}_i\in \mathbb{R}^{q\times q}\ |\ i\in [t-m,t], i\in \mathbb{Z} \}$ represents the input historical reanalysis 2D geopotential height data. Given the initial forecast time $t$ and the corresponding input data $\mathcal{X}_t$ and $\mathcal{G}_{t}$, the output multi-horizon TC track prediction can be computed by:
\begin{equation}
\mathcal{Y}_\tau = \mathbb{M}(\mathcal{X}_t,\mathcal{G}_t)
\end{equation}
where, $\mathcal{Y}_\tau=\{\textbf{Y}_{i}=(Lat_{t+i}-Lat_{t},Lon_{t+i}-Lon_t)\ |\ i\in[1,\tau], i\in \mathbb{Z}\}$, $Lat_t$ and $Lon_t$ are the Latitude and Longitude of TC center at time $t$. $\mathbb{M}(\cdot)$ represent the end-to-end DBF-Net. It should be noted that, we use the relative change in latitude and longitude $\textbf{Y}_{i}$ as the output of DBF-Net instead of the direct location. The reason for this is that the pressure field data is cropped from the center of TCs and the local information is more suitable for digging the relative changes in TCs motion.

\subsection{TC Features Encoder Module}
\begin{figure}[t]
\centering
\includegraphics[width=0.95\columnwidth]{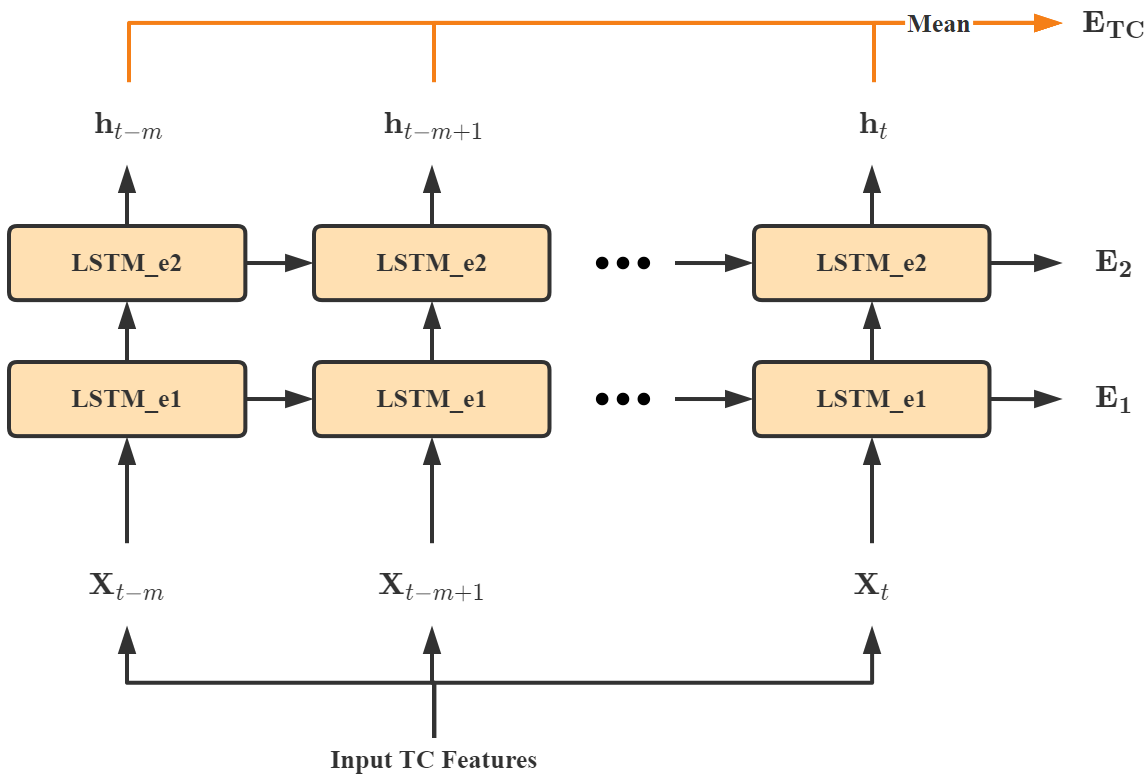} 
\caption{The LSTM-encoder module in TC Features Branch.}
\label{fig2}
\end{figure}
The TC features encoder module in the first branch of DBF-Net plays the role of encoding the inherent features of TCs $\textbf{X}_t=(x_1, x_2, x_3, x_4, x_5, x_6)$ at multiple historical times. Each $x_i$ in $\textbf{X}_t$ represents the latitude at time $t$, longitude at time $t$, maximum wind speed near the center at the bottom at time $t$, latitude difference between time $t$ and $t-1$, longitude difference between time $t$ and $t-1$ and wind speed difference between time $t$ and $t-4$ respectively. The features above are also the classical persistence factors in statistical forecasting methods. As a result, the purpose of the TC features encoder module is to encode time series features of the persistence factors.

As shown in Fig.2, the TC features encoder module consists of a two-layer stacked LSTM encoder. Each $\textbf{X}_t$ in sequence $\mathcal{X}_t$ is passed sequentially into the two-layer LSTM encoding module and produces the latent variable $\textbf{h}_t$.  The output of each LSTM encoding layer is fed into  the corresponding layer at next time step. The specific operation procedure in each LSTM encoding layer is as follows:
\begin{gather}
i_t=\sigma(W_{ii}\textbf{X}_t+b_{ii}+W_{hi}\textbf{h}_{t-1}+b_{hi})\\
f_t=\sigma(W_{if}\textbf{X}_t+b_{if}+W_{hf}\textbf{h}_{t-1}+b_{hf})\\
g_t=tanh(W_{ig}\textbf{X}_t+b_{ig}+W_{hg}\textbf{h}_{t-1}+b_{hg})\\
c_t =f_t\times c_{t-1}+i_t\times g_t\\
o_t=\sigma(W_{io}\textbf{X}_t+b_{io}+W_{ho}\textbf{h}_{t-1}+b_{ho})\\
\textbf{h}_t = o_t\times tanh(c_t)
\end{gather}
where the $i_t$, $f_t$ and $o_t$ represent the output of the input, forget and output gate in LSTM model. $W_{ii}(W_{hi})$, $W_{if}(W_{hf})$ and $W_{io}(W_{ho})$  are the correspond wight matrix related to the $\textbf{X}_t(\textbf{h}_{t-1})$. $c_t$ is the cell state that fed into the next time step together with the latent variable $\textbf{h}_t$. $\sigma(\cdot)$ is the Sigmoid function.

Given the input sequence $\mathcal{X}_t$ of length $m+1$ and the corresponding latent variable sequence $\{\textbf{h}_{t-m},\textbf{h}_{t-m+1},\cdots,\textbf{h}_{t}\}$, $t$ is the initial forecast time, we can compute the final time series code of the TC features by:
\begin{equation}
\textbf{E}_{\bf{TC}}= \frac{1}{m+1}\sum_{i=0}^{m}\textbf{h}_{t-i}
\end{equation}

\subsection{Pressure Field Branch}
In order to efficiently use the meteorological fields in the vicinity of TCs to improve the forecast accuracy. A 2D-CNN-based encoder-decoder networks are utilized to generate high-level spatio-temporal features from the reanalysis 2D geopotential height (GPH) data. 

\begin{figure}[t]
\centering
\includegraphics[width=0.8\columnwidth]{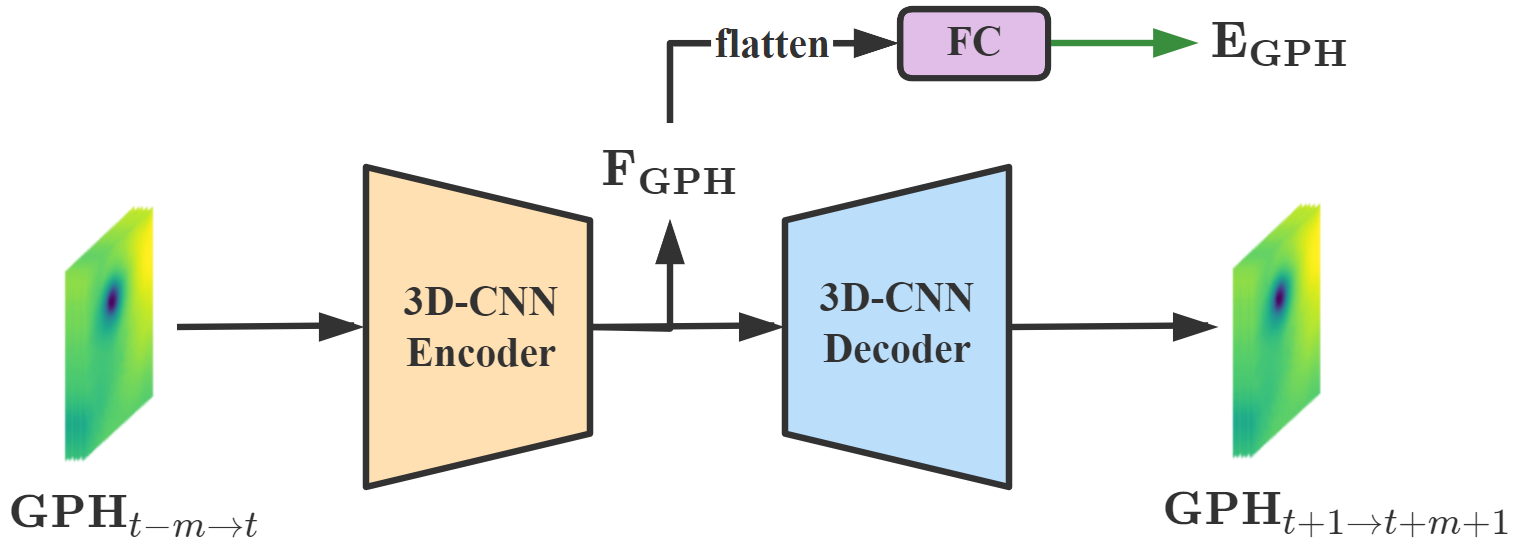} 
\caption{The pressure field branch architecture.}
\label{fig3}
\end{figure}

\begin{table}[h]
\centering
\caption{The 2D-CNN-based encoder architecture of the pressure field branch}
\begin{tabular}{lccc}
\hline
{\bf Layers}&{\bf Kernel Size}&{\bf Stride}&{\bf In Channel}\\\hline
{Conv\_1}&$3\times 3\times 3$&$1\times 1\times 1$&$1$\\
{MaxPool\_1}&$1\times 2\times 2$&$1\times 2\times 2$&$16$\\
{Conv\_2}&$3\times 3\times 3$&$1\times 1\times 1$&$16$\\
{MaxPool\_2}&$1\times 2\times 2$&$1\times 2\times 2$&$32$\\
{Conv\_3}&$3\times 3$&$1\times 1$&$32$\\
{MaxPool\_3}&$2\times 2$&$2\times 2$&$64$\\\hline
\end{tabular}
\end{table}

As shown in Fig. 3 and Table 1, the encoder of the pressure field branch contains three convolutional layers, the first two of which are 2D-CNN with kernel size $3\times 3\times 3$. To ensure that the GPH field data cover the full spatial extent that may affect the TC tracks, the window size at each historical time step $\textbf{GPH}_t$ of input 2D field $\mathcal{G}_t$ is set to $51\times 51$ values, which is approximately a radius of 1400km (the resolution of the reanalysis GPH data is 0.5 degrees). We choose LeakyReLU as the activation function of the encoder to enhance the nonlinear representation of the model. The output high-level spatio-temporal features can be computed by:
\begin{equation}
\begin{split}
\textbf{E}_{\textbf{GPH}}&=\textbf{FC}(\textbf{flatten}(\textbf{F}_{\textbf{GPH}}))\\
&=\textbf{FC}(\textbf{flatten}(\textbf{Encoder}(\mathcal{G}_t)))     
\end{split}
\end{equation}

where $\textbf{FC}(\cdot)$ is a fully connected layer. $\textbf{flatten}(\cdot)$ is the flatten operation that flattening output feature map of the 2D-CNN encoder.

As for the decoder in the pressure field branch, its structure is symmetrical with the 2D-CNN encoder and the transpose convolution is used to recover the spatio-temporal information from high-level features. It predicts the future $m+1$ time steps of the GPH, which is same length of time as the input $\mathcal{G}_t$. The loss function of  the pressure field branch is computed by:
\begin{equation}
\begin{split}
L_{GPH} &= \sum_{i=t+1}^{t+m+1}||\textbf{Decoder}(\textbf{F}_{\textbf{GPH}})-\textbf{TGPH}_i||_1\\
&=\sum_{i=t+1}^{t+m+1}||\textbf{GPH}_i-\textbf{TGPH}_i||_1  
\end{split}
\end{equation}
where $\textbf{TGPH}_i$ is the target value of the future GPH data. $||\cdot||_1$ is the $l_1$-norm function.

\subsection{Dual-Branched Features Fusion Decoder Module}

Based on the LSTM-based encoder in TC features branch and the pressure field branch mentioned above, there are three types of intermediate variables from two branches that fed into the LSTM-based decoder module, that is the final time series code of the TC features $\textbf{E}_{\textbf{TC}}$, the high-level spatio-temporal reanalysis 2D GPH features $\textbf{E}_{\textbf{GPH}}$ and the output features from LSTM encoder layers $\textbf{E}_1$ and $\textbf{E}_2$. In this subsection, we will introduce the LSTM-based decoder module for efficiently fusing the dual-branched multi-modal features and generating multi-horizon TC track forecasting results.

\begin{figure}[t]
\centering
\includegraphics[width=0.95\columnwidth]{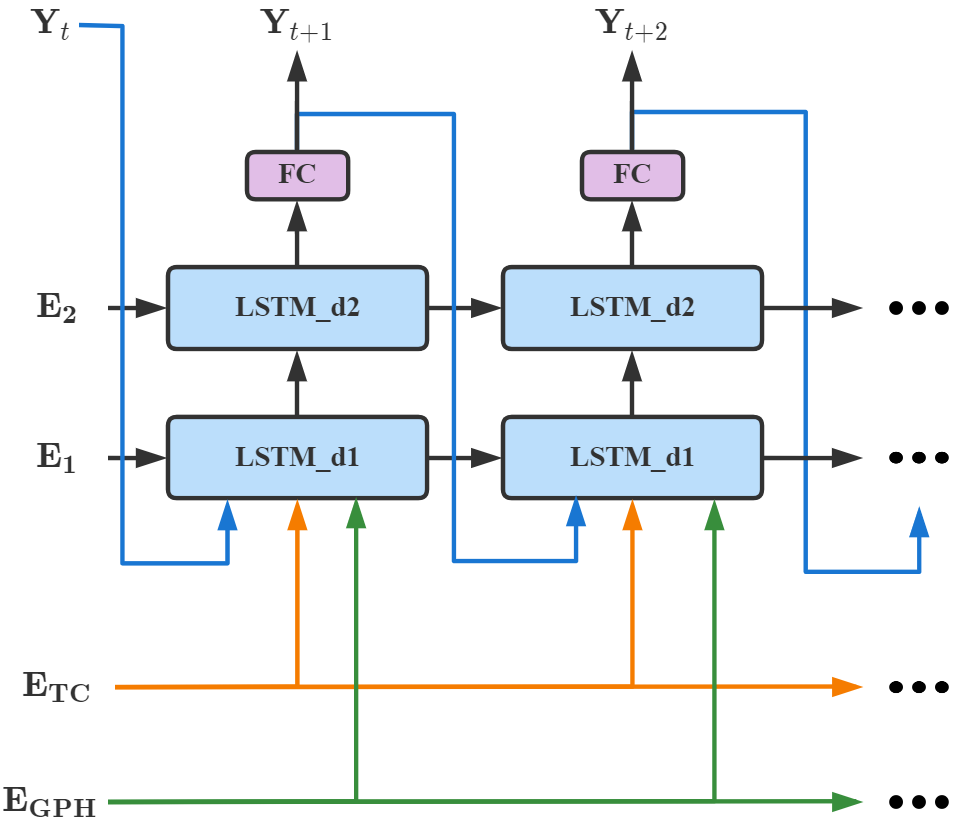} 
\caption{The LSTM-based decoder module in TC Features Branch with dual-branched features fusion.}
\label{fig4}
\end{figure}

Fig.3 illustrates the proposed LSTM-based decoder module that involves  a two-layers stacked LSTM decoder and the subsequent two fully connected layers. The specific decoding and feature fusion procedure is as follows:
\begin{gather}
\begin{split}
   i_t=\sigma(W_{Ti}\textbf{E}_{\textbf{TC}}&+W_{Gi}\textbf{E}_{\textbf{GPH}}+W_{Yi}\textbf{Y}_{t-1}\\&+W_{hi}\textbf{h}_{t-1}+b_i) 
\end{split}\\
\begin{split}
f_t=\sigma(W_{Tf}\textbf{E}_{\textbf{TC}}&+W_{Gf}\textbf{E}_{\textbf{GPH}}+W_{Yf}\textbf{Y}_{t-1}\\&+W_{hf}\textbf{h}_{t-1}+b_f)
\end{split}\\
\begin{split}
  g_t=(W_{Tg}\textbf{E}_{\textbf{TC}}&+W_{Gg}\textbf{E}_{\textbf{GPH}}+W_{Yg}\textbf{Y}_{t-1}\\&+W_{hg}\textbf{h}_{t-1}+b_g)  
\end{split}\\
c_t=f_t\times c_{t-1}+i_t\times g_t\\
\begin{split}
  o_t=\sigma(W_{To}\textbf{E}_{\textbf{TC}}+&W_{Go}\textbf{E}_{\textbf{GPH}}+W_{Yo}\textbf{Y}_{t-1}\\&+W_{ho}\textbf{h}_{t-1}+b_o)  
\end{split}\\
\textbf{h}_t=o_t\times tanh(c_t)\\
\textbf{Y}_t=\textbf{FC}_2(ReLU(\textbf{FC}_1(\textbf{h}_t)))
\end{gather}

where the initial state $\textbf{h}_t$ and $c_t$ are the elements of $\textbf{E}_1$ and $\textbf{E}_2$, which are the final state of the LSTM encoder layers. The initial input of the LSTM decoder $\textbf{Y}_t$ is set to zero.  The loss function of the LSTM-based decoder is also $l_1$-norm function, that is:
\begin{equation}
\begin{split}
L_{loc} &= \sum_{i=t+1} ^{t+\tau}||\textbf{Decoder}(\textbf{E}_{\textbf{TC}},\textbf{E}_{\textbf{GPH}},\textbf{E}_1,\textbf{E}_2,\textbf{Y}_t)-\textbf{Y}_i||_1\\
&=\sum_{i=t+1} ^{t+\tau}||\textbf{T}_i-\textbf{Y}_i||_1
\end{split}
\end{equation}
where, $\textbf{T}_i$ is the ground-truth changes of latitude and longitude. With the operation above, the features from both inherent TC features and reanalysis 2D pressure field can be fused effectively and we can achieve the multi-horizon TC track forecasting results $\mathcal{Y}_t$ based on the fused multi-modal features.
\subsection{Algorithmic Details}
We trained our proposed DBF-Net in a three stages manner. First, we only train the TC features encoder module by adding a fully connected layer to directly predict the target value of the TC track and get the pre-trained LSTM encoder in the TC features branch. Then, we utilize the reanalysis 2D pressure fields GPH data to train the pressure field branch of the DBF-Net and learn the temporal dynamic changing of GPH data. Finally, we add the LSTM decoder module into the training pipeline and train the DBF-Net in an end-to-end manner and the loss function at the final step is as follows:
\begin{equation}
L_{final} = L_{loc} + \alpha L_{GPH} + \beta L_2 
\end{equation}
where, the $L_2$ is the regularization term with $l_2$ penalty. $\alpha$ and $\beta$ are hyper-parameter. The training schedule detail will be discussed in Section 3.3. 
\section{Experiment}
\label{sec:typestyle}
In this section, we evaluate our proposed DBF-Net  on the best TC tracks data in Northwest Pacific (WNP). The good forecasting performance of DBF-Net is verified by the comparison with other deep-learning based and traditional TC track forecast methods. We also analyze the forecast results for several individual cases of TC events and the specific forecasting characteristics of the DBF-Net.

\subsection{Dataset}

{\bf Best track dataset (CMA-BST).} The inherent features data of TCs is extracted from the Best Track (BST) data released by China Meteorological Administration (CMA) \cite{TCdata}. It includes the location and intensity of TCs in the Northwest Pacific (WNP) Ocean ($0^{\circ}N\sim 50^{\circ}N$, $100^{\circ}E\sim 210^{\circ}E$) at six-hour intervals from 1949 to 2018. Examples of the CMA-BST data are shown in Table 2.

\begin{table}[h]
\centering
\caption{Examples of CMA-BST data. {\bf I} stands for  the intensity level of TCs. {\bf LAT} and {\bf LON} are the latitude and longitude of TCs' centers (unit: $\times 0.1^{\circ}$). {\bf PRES} stands for the central minimum pressure (unit: hPa). {\bf WND} and {\bf OWD} stand for the 2-minute maximum and average near-center wind speed respectively (unit: m/s).}
\setlength{\tabcolsep}{1mm}{
\begin{tabular}{ccccccc}
\hline
{\bf YYYYMMDDHH}&{\bf I}&{\bf LAT}&{\bf LON}&{\bf PRES}&{\bf WND}&{\bf OWD}\\\hline
1953061506&0&125&1116&1000&10&15\\
1953061512&0&132&1117&1000&10&15\\
1953061518&0&142&1117&1000&10&15\\
1953061600&0&150&1117&1000&10&20\\
1953061606&0&159&1112&999&10&20\\\hline
\end{tabular}}
\end{table}

\noindent{\bf Geopotential Height dataset (CFSR-GPH).} The reanalysis 2D geopotential height (GPH) data in pressure field branch is collected from Climate Forecast System Reanalysis (CFSR) dataset released by The National Centers for Environmental Prediction (NCEP) \cite{NCEP} . CFSR-GPH is grid data with a spatial resolution of $0.5^{\circ}$ and the temporal resolution is aligned with CMA-BST from 1979 to present. TCs in WNP are mostly genrated at the southern edge of the subtropical high pressure and move along its periphery. Therefore, the 500hPa geopentential height data is chosen as the background pressure field to describe the activity of TCs.

\noindent{\bf Dataset Split}. Based on the CMA-BST and CFSR-GPH dataset mentioned above. We choose overlap of the two datasets i.e. TCs from 1979 to 2018. And we only keep the TCs with a life cycle greater than four days to ensure the persistence. There are 940 TCs left in this dataset. We make 17000+ samples for model training, validating and testing based on a sliding window of length \emph{input sequence length} $+$ \emph{prediction length} (as shown in Table 3).
\begin{table}[h]
\centering
\caption{Dataset splitting based on CMA-BST and CFSR-GPH.}
\setlength{\tabcolsep}{2mm}{
\begin{tabular}{cccc}
\hline
&{\bf Train Set}&{\bf Val Set}&{\bf Test Set}\\\hline
Years&1979$\sim$2008&2009$\sim$2013& 2014$\sim$2018\\
\#TCs&723&104&113\\
\#Samples&13233&1847&2174\\\hline
\end{tabular}}
\end{table}

\subsection{Metrics}
In order to evaluate the TCs track forecast results, the Mean Distance Error (MDE) is the common metrics to measure the average distance error between model prediction and ground truth. MDE can be computed by:
\begin{equation}\small
\begin{split}
MDE &= 2\times R\times \arcsin\\
&\sqrt{\sin^2(\frac{\varphi_{pre}-\varphi_{gt}}{2})+\cos \varphi_{pre}\cos\varphi_{gt}\sin^2(\frac{\lambda_{pre}-\lambda_{gt}}{2})}\\
&\approx \sqrt{\Delta Lat^2+\Delta Lon^2}\times 110
\end{split}
\end{equation}
where, $R\approx 6371km$ represents the radius of earth. $\varphi_{pre}$ and $\varphi_{gt}$ stand for the latitude value of prediction and ground truth. $\lambda_{pre}$ and $\lambda_{gt}$ stand for the longitude value of prediction and ground truth. 

Besides, the skill score is also the index to evaluate the practical availability of the methods, as follow:
\begin{equation}
    skill\ score=\frac{e_A-e_B}{e_A}\times 100\%
\end{equation}
where, $e_A$ is the prediction error of CLIPER method and $e_B$ is the error of proposed method.

\subsection{Implementation Details}
We train our proposed DBF-Net in a three stages manner with the Pytorch framework, which has been discussed in Section 2.5. We use the RMSProp optimizer and set the initial learning rate to 0.001. The batch size of training set is set to 64. The hyper parameter $\alpha$ and $\beta$ in equation (19) is set to 1.2 and 0.00001 respectively.

For multi-horizon forecasting (i.e. predicting the TCs track at multiple future time steps), the output prediction sequence length of the DBF-Net is 4 and the input sequence length is 5. That is we predict the 6h, 12h, 18h and 24h TCs tracks based on the historical data from time $t-5$ (30h prior) to time $t$ (the current time). We train our DBF-Net on a single NVIDIA GeForce GTX 3090 GPU.

\subsection{Comparison with Statistical/Deep Learning Forecast Methods}
We first compare our proposed DBF-Net with other statistical and deep learning based TCs track forecast methods, including the extrapolation method, CLIPER method, feature fusion network \cite{2019Tropical} and recent BiGRU-attn \cite{BiGRU-attn}. The extrapolation  is a simple traditional TCs track forecast method. It assumes that the direction and speed of TCs movement do not change much, and predicts based on the movement direction and velocity at previous times. CLIPER can be treated as the benchmark of other track forecast methods. It uses correlation analysis to screen climate persistence factors and constructed multivariate linear regression models. In this paper,  we replace the multivariate linear model with a back propagation (BP) neural network model, which enhancing the nonlinear representation of the CLIPER model. We selected 20 factors with strong correlation from 46 climate persistence factors by Pearson correlation analysis and feed them into the BP neural network.

\begin{table}[h]
\centering
\caption{Comparison of the TCs track forecasting results of statistical and deep learning methods.}
\renewcommand\tabcolsep{6pt}
\resizebox{0.95\columnwidth}{!}{
\begin{tabular}{ccccc}
\toprule
\multirow{2}{*}{\bf Methods}&\multicolumn{4}{c}{\bf MDE (km)}\\
\cmidrule{2-5}
&{6h}&{12h}&{18h}&24h\\
\midrule
extrapolation&33.78&79.20&135.48&201.28\\
CLIPER-BP&37.53&73.31&115.13&162.62\\
FFN \cite{2019Tropical}&32.90&-&-&136.10\\
BiGRU-attn \cite{BiGRU-attn}&-&-&-&147.38\\
DBF-Net (Ours)&{\bf 31.30}&{\bf 58.94}&{\bf 87.60}&{\bf 119.05}\\
\bottomrule
\end{tabular}}
\end{table}

As shown in Table 4, our proposed DBF-Net outperforms previous works. Specifically, compared with the benchmark forecast method CLIPER, DBF-Net achieves better MDEs for all forecast time steps. That is the skill score with respect to the CLIPER is positive, which demonstrates the practical availability of our method. In addition, our DBF-Net also ourperforms previous deep learning based methods. Compared with FFN \cite{2019Tropical} that utilizes the wind, pressure fields simultaneously, our DBF-Net achieves better results only based on pressure field. This also shows that our method can better encode the effective features of the input data.

\subsection{Comparison with NWP Forecast Methods}
We further compare our DBF-Net with the Numerical Weather Prediction (NWP) system that commonly used in operational forecasting. The global pattern T213/T639 and Shanghai typhoon region pattern (SHTP) are chosen for comparison. Compared with our deep learning based method, the NWP methods always need a great number of computation resources and the inference time increases rapidly as input data resolution increases. However, NWP methods still can achieve better forecast accuracy compared with deep learning based methods. As shown in Table 5, DBF-Net could achieve comparable performance compared with global pattern T213/T639, especially in year 2014 and 2015, the 24h MDE of DBF-Net is much better than T213/T639. However, compared with the region pattern SHTP, the forecast of the DBF-Net still has a certain gap. The great performance of the SHTP may due to the high-resolution multi-layer nested grid input data and huge computational resource consumption. In contrast, our proposed DBF-Net achieves relatively high prediction accuracy under the premise of low-resolution input (1$^{\circ}$ spatio resolution for GPH data) and small computational resource consumption. At the same time, we believe that by using higher resolution data for model training, our DBF-Net could further improve the forecast accuracy which can be further studied in the future works.

\begin{table}[h]
\centering
\caption{Comparison of the TCs track forecasting results of NWP methods. The results of NWP methods is released by [xx,xx,xx,xx].}
\resizebox{0.95\columnwidth}{!}{
\begin{tabular}{ccccc}
\toprule
\multicolumn{2}{l}{\bf Year}&{\bf T213/T639}&{\bf SHTP}& {\bf DBF-Net}\\
\midrule
\multirow{2}{*}{2014}&\#Samples&411&112&330\\
&24h MDE (km)&121.8&{\bf 64.9}&107.4\\
\midrule
\multirow{2}{*}{2015}&\#Samples&46&440&671\\
&24h MDE (km)&150.6&{\bf 67.8}&107.0\\
\midrule
\multirow{2}{*}{2016}&\#Samples&412&194&332\\
&24h MDE (km)&114.9&{\bf 88.5}&118.9\\
\midrule
\multirow{2}{*}{2017}&\#Samples&301&253&319\\
&24h MDE (km)&98.7&{\bf 89.1}&130.6\\
\bottomrule
\end{tabular}}
\end{table}

\subsection{Ablation Study} 
In order to verify the effectiveness of our proposed Dual-Branched spatio-tempotal Fusion Network (DBF-Net) architecture, we experimented with different branching structures. As can be seen in Table 6, our proposed DBF-Net with two branches encoder-decoder networks and feature fusion module, which is denoted as "DBF-Net$\star$", achieves the best forecast MDE except with the 6h forecast result in "DBF-Net" that verified the consistency of our method. In addition, the relatively bad performance of LSTM-based and 2D-CNN-based encoder-decoder architecture alone, which is denoted as "TC-features-only" and "Pressure-fields-only", demonstrate that it is hard to obtain good forecast results for single inherent features or meteorological fields input. It makes sense to fuse these two types of data for better forecast accuracy. The results in Table 6 also show the effectiveness of the 2D-CNN Decoder module for enhancing the temporal dynamic changing of GPH data.

\begin{table}[h]
\centering
\caption{The impact of different branches in DBF-Net. $\star$ represents the model with 2D-CNN Decoder in pressure branch to enhance the temporal dynamic changing of GPH data.}
\renewcommand\tabcolsep{6pt}
\resizebox{0.95\columnwidth}{!}{
\begin{tabular}{ccccc}
\toprule
\multirow{2}{*}{\bf Architecture}&\multicolumn{4}{c}{\bf MDE (km)}\\
\cmidrule{2-5}
&{6h}&{12h}&{18h}&24h\\
\midrule
TC-features-only&32.87&69.08&111.08&158.11\\
Pressure-fields-only&34.35&64.07&95.83&130.79\\
Pressure-fields-only$\star$&34.27&63.31&93.63&126.89\\
DBF-Net&{\bf 31.18}&59.13&88.82&121.79\\
DBF-Net$\star$& 31.30&{\bf 58.94}&{\bf 87.60}&{\bf 119.05}\\
\bottomrule
\end{tabular}}
\end{table}

It should be noted that the 2D-CNN decoder module in pressure fields branch only plays a role in model training. Once the DBF-Net is trained, the inference is done by just passing TC features branch and 2D-CNN encoder module. Therefore, the 2D-CNN decoder module does not increase the memory and computational cost of the DBF-Net inference.


\subsection{Case Study}

In this subsection, we select three individual cases of TC events, namely Typhoon Trami 1824, Typhoon Hagibis 1919 and Typhoon Fengshen 1925. According to the forecast results, the validity of the DBF-Net is further verified, and the forecast characteristics of DBF-Net are analyzed.

\begin{figure}[t]
\centering
\includegraphics[width=0.95\columnwidth]{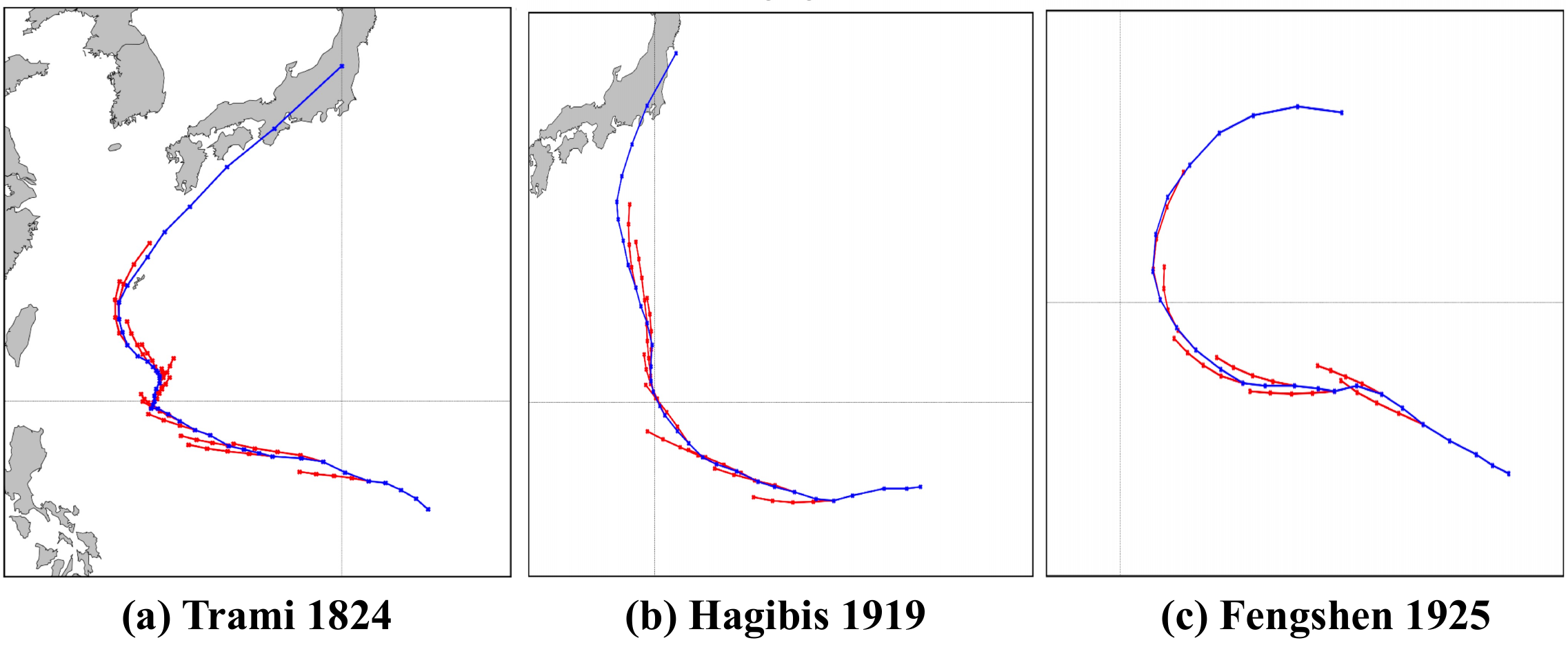} 
\caption{Exmaple of TCs track forecast results. The blue line represents the ground truth track. The red line represents the forecast results.}
\label{fig5}
\end{figure}

Fig. 5 and Table 7$\sim$9 shows the TCs track forecast results for the three cases. In Fig.5, the blue line represents the ground truth track of TCs and the red line is the forecast results of the proposed DBF-Net. The intersection of the blue and red lines is the location of initial forecast time. The 4 points extending from the red line represent the forecast path in the next 24 hours (6-hour interval) from the initial forecast time. 

\begin{table}[h]
\centering
\caption{The track forecast result of DBF-Net for Typhoon Trami 1824. {\bf INT} stands for the intensity level of the TC. {\bf AVG} stands for the average prediction MDE.}
\renewcommand\tabcolsep{6pt}
\resizebox{0.95\columnwidth}{!}{
\begin{tabular}{cccccc}
\toprule
\multicolumn{6}{c}{\bf Typhoon Trami 1824}\\
\midrule
\multirow{2}{*}{\bf MMDDHH}&\multirow{2}{*}{\bf INT}&\multicolumn{4}{c}{\bf MDE (km)}\\
\cmidrule{3-6}
&&{6h}&{12h}&{18h}&24h\\
\midrule
092118&TS&54.56&111.16&137.78&194.56\\
092206&STS&19.60&41.45&40.61&75.08\\
092218&TY&63.42&103.65&136.77&147.31\\
092306&STY&5.89&49.85&62.80&92.46\\
092318&STY&1.05&26.49&46.63&68.26\\
092406&SuperTY&11.45&21.48&11.23&66.31\\
092418&SuperTY&22.59&33.89&65.07&83.58\\
092506&SuperTY&8.75&23.03&34.63&20.56\\
092518&SuperTY&24.11&27.85&52.71&62.20\\
092606&STY&23.19&51.10&83.84&129.74\\
092618&STY&26.29&44.59&65.90&94.28\\
092706&STY&13.92&28.89&62.13&86.24\\
092718&STY&33.67&58.95&55.01&50.70\\
092806&STY&18.94&24.64&26.86&52.50\\
092818&STY&3.85&23.75&91.57&108.31\\
\midrule
\multicolumn{2}{c}{\bf AVG}&23.48&41.70&58.40&82.43\\
\bottomrule
\end{tabular}}
\end{table}

\begin{table}[h]
\centering
\caption{The track forecast result of DBF-Net for Typhoon Hagibis 1919. {\bf INT} stands for the intensity level of the TC. {\bf AVG} stands for the average prediction MDE.}
\renewcommand\tabcolsep{6pt}
\resizebox{0.95\columnwidth}{!}{
\begin{tabular}{cccccc}
\toprule
\multicolumn{6}{c}{\bf Typhoon Hagibis 1919}\\
\midrule
\multirow{2}{*}{\bf MMDDHH}&\multirow{2}{*}{\bf INT}&\multicolumn{4}{c}{\bf MDE (km)}\\
\cmidrule{3-6}
&&{6h}&{12h}&{18h}&24h\\
\midrule
100606&STS&28.26&68.36&90.68&107.11\\
100618&TY&11.00&30.92&31.60&30.41\\
100706&SuperTY&38.75&62.95&26.88&41.82\\
100718&SuperTY&42.27&75.44&132.27&177.07\\
100806&SuperTY&26.80&24.93&60.36&77.02\\
100818&SuperTY&51.18&54.16&82.65&93.71\\
100906&SuperTY&30.06&56.08&45.70&9.55\\
100918&SuperTY&29.00&56.08&86.27&109.05\\
101006&SuperTY&47.06&75.26&90.90&99.15\\
101018&STY&28.45&52.75&84.87&97.22\\
\midrule
\multicolumn{2}{c}{\bf AVG}&29.43&51.41&66.84&86.77\\
\bottomrule
\end{tabular}}
\end{table}

\begin{table}[h]
\centering
\caption{The track forecast result of DBF-Net for Typhoon Fengshen 1925. {\bf INT} stands for the intensity level of the TC. {\bf AVG} stands for the average prediction MDE.}
\renewcommand\tabcolsep{6pt}
\resizebox{0.90\columnwidth}{!}{
\begin{tabular}{cccccc}
\toprule
\multicolumn{6}{c}{\bf Typhoon Fengshen 1925}\\
\midrule
\multirow{2}{*}{\bf MMDDHH}&\multirow{2}{*}{\bf INT}&\multicolumn{4}{c}{\bf MDE (km)}\\
\cmidrule{3-6}
&&{6h}&{12h}&{18h}&24h\\
\midrule
111206&TS&34.07&50.98&26.22&63.83\\
111218&TS&39.65&99.83&116.08&183.11\\
111306&STS&44.77&38.92&48.00&60.31\\
111318&STS&59.19&75.96&90.71&152.12\\
111406&STS&27.19&81.96&126.19&183.04\\
111418&TY&14.69&68.90&102.61&140.09\\
111506&STY&9.65&17.76&38.13&50.84\\
\midrule
\multicolumn{2}{c}{\bf AVG}&28.05&62.49&92.39&137.03\\
\bottomrule
\end{tabular}}
\end{table}

For Typhoon Trami 1824, its path generally shows a trend of first westward and then northward. As shown in Fig. 5(a), the error between the forecast and the ground truth track is relatively small at the inflection point from west to north. This shows that the model itself has learned the potential features of TCs track movement. For Typhoon Hagibis 1919, there also a inflection point at time "100718" (as shown in Fig.5(b) and Table 8). Although the forecast result is relatively bad (177.07km for 24h forecasting), the DBF-Net also could fix the prediction by bringing the observations from the next time step (24.93km for 12h forecasting at time "100806"). This shows that the historical information closest to the initial forecast time is more important. For Typhoon Fengshen 1925, its trajectory presents a 180$^{\circ}$ turning trend, which is unconventional. As shown in Fig.5(c), DBF-Net could correctly predict the turning trend of the TC. However, the forecasting length of the track vector is uniformly smaller than the ground truth and causes the average MDE to be relatively large (in Table 9). 

In Table 7$\sim$9, we also report the intensity level (INT) at each time step and classify it into 6 categories, namely tropical depression (TD), tropical storm (TS), severe tropical storm (STS), typhoon (TY), severe typhoon (STY) and super severe typhoon (SuperTY). By comparing the results in Tables 7$\sim$9, it can be found that the DBF-Net has a relatively lower MDE when the intensity level is larger. Especially for the TCs with the intensity level of "SuperTC", the predicted MDE is comparable to the NWP model. This phenomenon also explains the relatively poor forecast results of Typhoon Fengshen 1925.

\section{Conclusion}
\label{sec:majhead}

In this paper, we explore the way to forecast the TCs track for multiple future time steps by proposing a novel deep learning based model, named DBF-Net, to make full use of both inherent features of TCs and reanalysis 2D pressure fields data and fuse the multi-modal features efficiently. DBF-Net contains two branches with encoder-decoder architectures and can be split into three part. The first part is the LSTM-based TC features encoder module that captures the high-level temporal features from historical inherent features of TCs. The second part is the pressure field branch that extracts the spatio-temporal features by learning the temporal dynamic changing of GPH data with a 2D-CNN-based encoder-decoder network. The last part is a LSTM-based decoder module that fuses the  multi-modal high-level features from different branches and produces the multi-horizon prediction. The experiments on the TCs track dataset in the Northwest Pacific Ocean verify the effectiveness of our proposed DBF-Net.

\section{Acknowledge}
\label{ssec:subhead}

This paper is supported by.




\bibliographystyle{IEEEbib}
\bibliography{test}


\end{document}